# Design, Construction and Implementation of Stewart Platform – Control of Rolling Ball on Platform Through Artificial Vision

Alejandro Maldonado, Cristian Bueno, David Loza, Alexander Ibarra.

Energy and Mechanical Department - Mechatronic Engineering, Universidad de las Fuerzas Armadas-ESPE.

**Abstract**

Artificial vision (AV) has recently emerged as an extremely important tool to help control robots with or without minimal human interaction. This article presents the design and construction of a parallel robot called a Stewart platform. Using Python as the main programming language, we implement an AV module with modern feedback control techniques that guide the position of a rolling ball over a Stewart platform.

## I.    INTRODUCTION

A parallel robot consists of a mechanism with two bases (upper and bottom) connected to each other through at least two linked bars or kinematic chains. Several implementations of such parallel manipulators with five or six degrees of freedom exist. Due to this relatively high number of degrees of freedom parallel robots tend to have a smaller workspace than serial robots.

Mobility is the most common feature used to classify parallel robots, leading to two distinct categories: planar and spatial mobility. The Stewart-Gough platform, invented in 1965 by D. Stewart from Institute of Mechanical Engineers in the United Kingdom, falls under the spatial mobility category due to its capacity to move about the X, Y, and Z axes (Fig. 1). In the past century, the Stewart platform configuration has been one of the most commonly used in the context of parallel robots with spatial mobility.

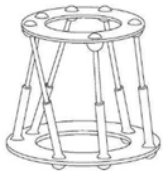

**Fig. 1: Sketch of Stewart Platform with 6 degrees of freedom**

The Stewart Platform is configured with 6 degrees of freedom (DOF), which allow the robot to use longitudinal displacement (linear motion in X, Y, and Z axes), as well rotational displacement (pitch, yaw, and roll). One of the approaches used to solve the kinematics of the Stewart platform is to analyze its two platforms (also referred to as bases) and their corresponding actuators (upper and bottom) separately. In particular, the bottom platform is taken as the fixed base, while the upper platform is taken as the mobile base. The six DOF are determined by the actuators. It has a high load-to-power relationship due to its drive-train, which is directly connected between the upper and lower bases, thus allowing it to lift objects significantly heavier than itself. Another benefit, as a result of its configuration, is its ability to reach high speeds and accelerations, making it a highly efficient robot. There exist many applications for the Stewart platform, including dynamic ocean wave simulators, CNC machines, stabilizing platforms capable of eliminating unwanted external movement, positioning systems for parabolic satellite antennas, and "pick and place" robots for assembly tasks, among others.

In order to automatically control the Stewart platform, kinematics, dynamics and AV recognition modules are essential. Here, using Python, we implement a fuzzy control system with an AV module to automatically collect positional information for a ball rolling over a Stewart platform. The AV module recognizes the platform and the rolling ball, obtaining centroids, limits, and positions in real time. To know and select which motor meets the specifications for this project, we use the Lagrange-Euler method together with vector operations that help us to obtain its parameters (torque, operation angle, etc) and analyze the kinematics and dynamics of the platform.

## II.    KINEMATICS AND DYNAMICS

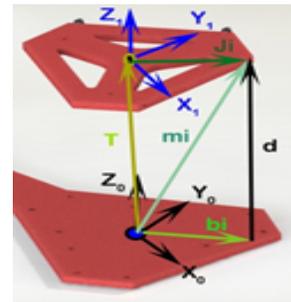

**Fig. 2: Labelled diagram of the Stewart platform with the main variables.**

- $X_0, Y_0, Z_0$ = coordinates for the manipulator fixed reference frame;
- $X_1, Y_1, Z_1$ = coordinates for the manipulator mobile reference frame;
- i: beam index (1-6);
- T = translation vector;
- mi = motion vector of the upper joint/fixed system relation;
- $d_i$ = distance between the platform and the base;
- $bi_i$ = Coordinates of the actuator in the fixed reference frame.
- $J_i$ = Coordinates of the joint in the mobile reference frame.

In order to get the $m_i$ vector coordinates with respect to the fixed reference frame, it is necessary to analyze the following equation:

$$\vec{T} + \vec{Ji} = \vec{mi}$$

We use the rotational matrix of the system ($R_t$), which is the product of each single-axis rotational matrix. Having obtained $R_t$ we can calculate Ji:

$$\vec{Ji} = \vec{Rt} \cdot \vec{J}$$

Replacing the mobile coordinates of the platform (2) into the equation for $m_i$ (1), we obtain:

$$\vec{mi} = \vec{T} + \vec{Rt} \cdot \vec{J}$$

The vector $mi$ can also be written as:

$$\vec{mi} = \vec{bi} + \vec{di}$$

Using the last two equations, we obtain:

$$\vec{di} = \vec{T} + \vec{Rt} \cdot \vec{J} - \vec{bi}$$

This equation gives us an expression for the length of separation between the upper platform and the actuators in the bottom base, which is essential for the study of the platform's kinematics.

We obtain the platform's dynamics using the Lagrange-Euler method. With this approach, we can use the geometric parameters and inertias of each joint in the platform.

$$\frac{d}{dt}\left(\frac{\partial L}{\partial \dot{q}i}\right) - \left(\frac{\partial L}{\partial qi}\right) = \tau_i \qquad i = 1, 2, 3, \ldots, n$$

Where:

- n = number of the joints in the manipulator.

- L = K-P = system Lagrangian;
- K = kinetic energy;
- P = potential energy;
- $q_i$ = generalized coordinates for the manipulator joints;
- $\dot{q}_i$ = derivative of $q_i$ respect Time;
- $\tau$ = applied force over the joint to produce motion.

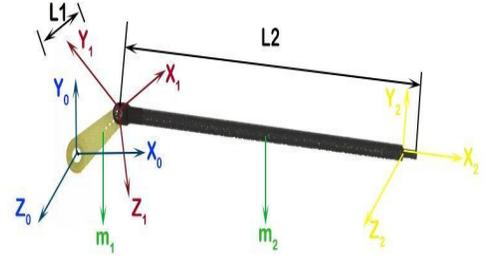

**Fig. 3: Schematic diagram for a single joint in the Stewart platform and its associated parameters required to calculate its kinematics and dynamics**

We define the homogeneous matrix of the joint as:

$$^0A_2 = {^0A_1} \times {^1A_2}$$

and use the rotation matrix (Q) and equation (9) to analyze the movement of each linked bar:

$$Q = \begin{bmatrix} 0 & -1 & 0 & 0 \\ 1 & 0 & 0 & 0 \\ 0 & 0 & 0 & 0 \\ 0 & 0 & 0 & 0 \end{bmatrix}$$

$$U_{ij} = \left[Q_j^{j-1} A_i\right]; \quad when\ j \leq i$$

$$U_{ij} = [0]; \quad when\ j \geq i$$

Thus the U matrixes are given by:

$$U_{01} = Q_1^0 A_1$$

$$U_{12} = Q_1^0 A_2$$

$$U_{02} = \, ^0A_1 Q_2^1 A_2$$

Each linked bar's inertia is obtained using the inertia matrix:

$$J_i = \begin{bmatrix} \frac{-I_{xx}+I_{yy}+I_{zz}}{2} & I_{xy} & I_{xz} & m_i\overline{x}_i \\ I_{xy} & \frac{I_{xx}-I_{yy}+I_{zz}}{2} & I_{yz} & m_i\overline{y}_i \\ I_{xz} & I_{yz} & \frac{I_{xx}+I_{yy}-I_{zz}}{2} & m_i\overline{z}_i \\ m_i\overline{x}_i & m_i\overline{y}_i & m_i\overline{z}_i & m_i \end{bmatrix}$$

when $i = 1, 2, 3, ....., n$

Kinetic matrices are expressed as follows:

$$D_{11} = Tr(U_{01} \cdot J_1 \cdot U_{01}^T) + Tr(U_{12} \cdot J_1 \cdot U_{12}^T)$$

$$D_{21} = D_{12} = Tr(U_{02} \cdot J_2 \cdot U_{12}^T)$$

$$D_{22} = Tr(U_{22} \cdot J_2 \cdot U_{22}^T)$$

Taking into consideration the Coriolis and centrifugal forces from each link bar, we perform a deep Lagrange-Euler dynamic analysis:

$$h_1 = \sum_{k=1}^{2}\sum_{m=1}^{2} h_{1km}\dot{\theta}_k\dot{\theta}_m = h_{111}\dot{\theta}_1\dot{\theta}_1 + h_{112}\dot{\theta}_1\dot{\theta}_2 + h_{121}\dot{\theta}_2\dot{\theta}_1 + h_{12}$$

$$h_2 = \sum_{k=1}^{2}\sum_{m=1}^{2} h_{2km}\dot{\theta}_k\dot{\theta}_m = h_{211}\dot{\theta}_1\dot{\theta}_1 + h_{212}\dot{\theta}_1\dot{\theta}_2 + h_{221}\dot{\theta}_2\dot{\theta}_1 + h_{22}$$

Where $h_1$ and $h_2$ represent the forces of the servo arm and link bar attached to it respectively.

The potential energy can be expressed as:

$$C_1 = -\left(m_1 \cdot g \cdot U_{01} \bullet .\overline{1r_1} + m_2 \cdot g \cdot U_{12} \bullet .\overline{2r_2}\right)$$

$$C_2 = -\left(m_2 \cdot g \cdot U_{02} \bullet .\overline{2r_2}\right)$$

where:

- $nr_n$: coordinate matrix of the platform's link bar (with dimensions 4x1).
- $g$: gravity matrix (with dimensions 1x4).
- $m_1$ and $m_2$: Link bars' mass.

Hence, we have obtained equations for each energy term (kinetic, Coriolis, potential and centrifugal). We can then substitute them into the Lagrange-Euler equation, which gives us the torque for each joint in the robot.

$$\tau = D(\theta)\ddot{\theta}(t) + h(\theta, \dot{\theta}) + c(\theta)$$

### III. ANGLE OF THE SERVOMOTOR

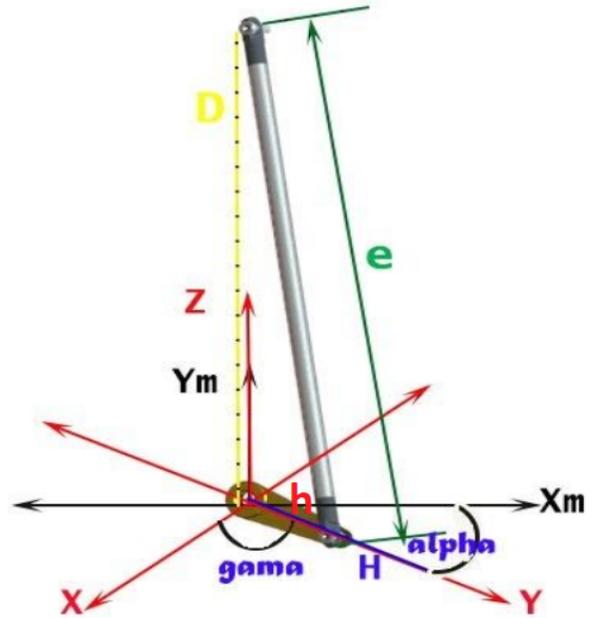

**Figure 4: Parameters in a joint required to calculate the angle of the servomotor.**

The diagram above shows the parameters of a joint that are required to obtain the angle of the servomotor. These parameters are:

- h = Length of the servomotor's arm over its own axle (different from Coriolis and centrifugal force parameter).
- H = coordinate vector of the servomotor's arm.
- e = length of the link bar between the upper platform and the servomotor.
- Di = distance (not fixed) between the servomotor's arm and the upper platform joint.
- α (alpha) = angle between the servomotor's arm and its own x-axis.

- ɣ (gamma) = angle between the servomotor arm and the platform frame of reference.

Using basic trigonometric relations, α can be expressed as:

$$\alpha = \left(\frac{d^2 - e^2 + h^2}{\sqrt{a^2 + b^2}}\right) - \left(\frac{b}{a}\right)$$

where "a" and "b' are defined by the following equations:

$$a = 2h \times (Z_{J_i} - Z_{B_i})$$

$$b = 2h \times (\cos(\lambda) \times \left(X_{J_i} - X_{B_i}\right) + \sin(\lambda) \cdot \left(\left(X_{J_i} - X_{B_i}\right)\right))$$

Once α is thus obtained for each position, it is then possible to control the servo motor through pulse width modulation (PWM).

## IV. ARTIFICIAL VISION

Next, we built an artificial vision module in Python to recognize the platform's surface and the rolling ball. A Gaussian filter was used for surface identification, which allows us to filter out image noise and perform a deep scan of the work area. We then converted each frame to hue-saturation-value (HSV) format for better analysis.

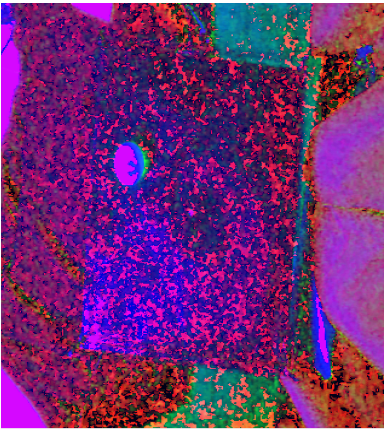

**Figure 5: Image of rolling ball on platform processed with a Gaussian filter and converted to HSV format.**

The region of interest was set using the bitwise Python function, and having obtained a clear image of the rolling ball on the platform, we transformed it to grayscale to decrease the image size and improve performance

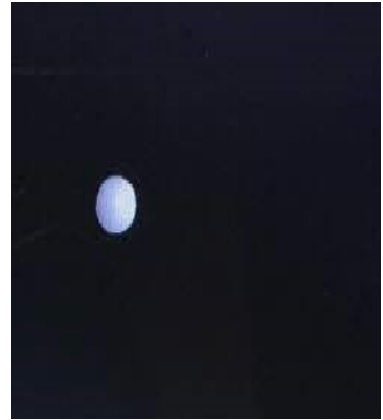

**Figure 6: Image transformed to grayscale.**

We then identified the ball and platform contours and edges, and with this information obtain their relative positions.

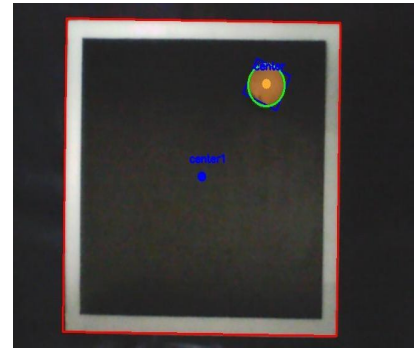

**Figure 6: Contour detection of the ball and platform.**

## V. METHOD

We combined a modern control technique, Fuzzy control, with our artificial vision module. The main advantage of this control method is that the explicit equations of motion for the system are no longer necessary, allowing us to avoid the complex process of solving for the equations that represent the Stewart platform. Fuzzy control technique uses a set of rules (functions) that describe the system's behaviour when the ball rolls in the plane of motion. For all our simulations, it is important to specify that the error signal is the distance between the centers of the platform and ball.

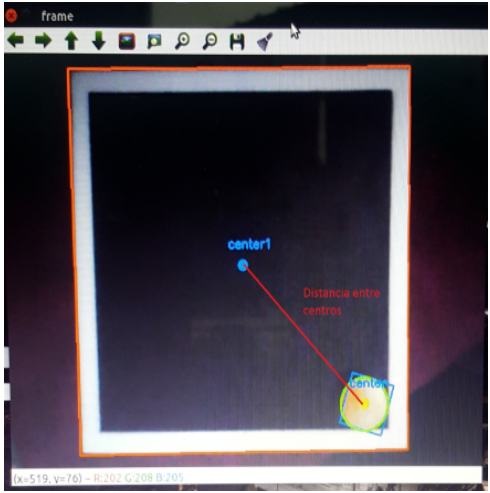

**Figure 7: Distance between centers.**

The derivative of this error signal is the rate of change of the ball's position and can be obtained from:

$$\frac{de}{dt} = \left(\frac{x(t_1) - x(t_2)}{t_1 - t_2}\right)$$

We have used different models with different approaches, inputs and outputs. These models are called Fuzzy 1, Fuzzy 2, Fuzzy 3 and Fuzzy PD.

- **Fuzzy 1**: As input values we used the position of the ball in X axis and Y axis and the Error signal. The set rules are defined as: ENG, ENP, EC, EPP, EPG.

| Inputs | Range in mm |
|---|---|
| Position in X axis | [-200, 200] |
| Position in Y axis | [-200, 200] |
| Error Signal | [-300, 300] |

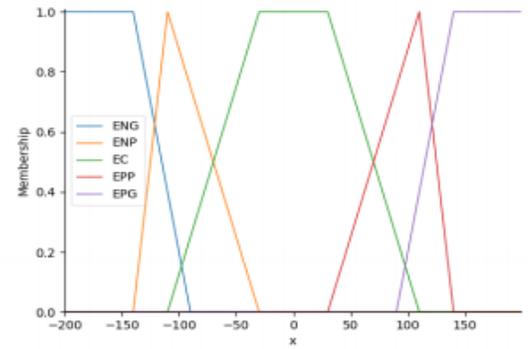
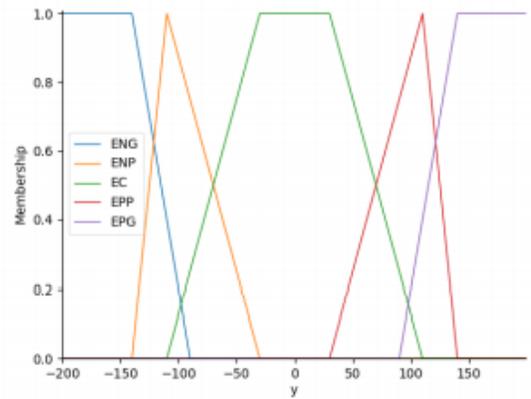
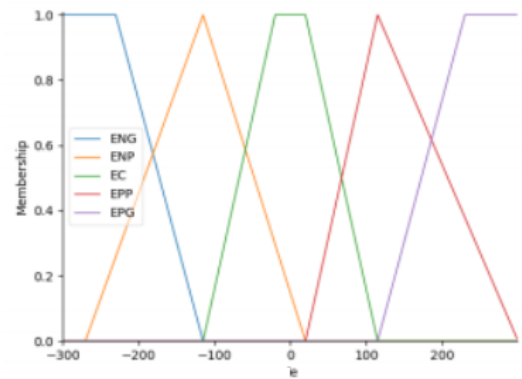

**Figure 8: Inputs with their Set of rules for Fuzzy 1.**

As outputs, Roll and Pitch parameters were considered and the set of rules are defined as: NG, NP, Z, PP, PG.

| Outputs | Range in degrees |
|---|---|
| Roll | [-6, 6] |
| Pitch | [-6, 6] |

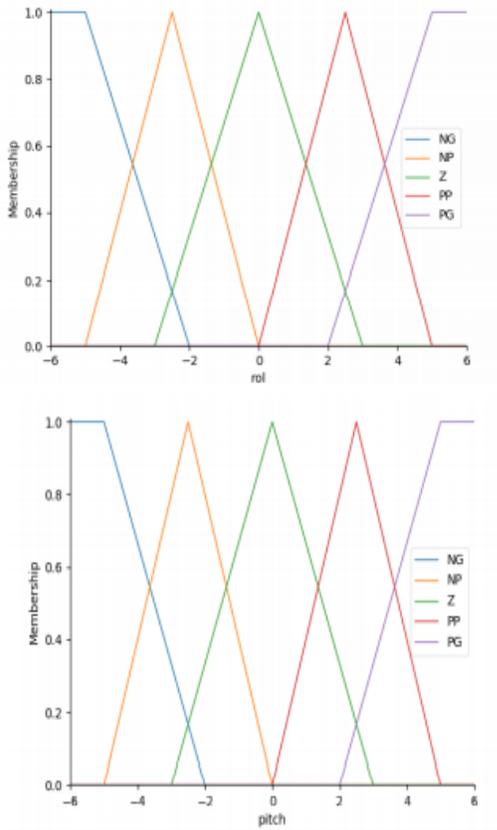

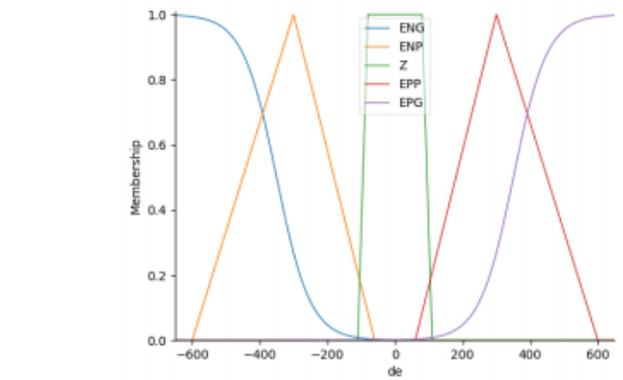

**Figure 10: Inputs with their Set of rules for Fuzzy 2.**

As output, only Roll parameter were considered and the set of rules are defined as: PNG, PNM, PNP, PC, PPP, PPM, PPG.

| Outputs | Range in degrees |
|---|---|
| Roll | [-6, 6] |

**Figure 9: Outputs with their Set of rules for Fuzzy 1.**

- **Fuzzy 2**: As input values we used the position of the ball in the X axis and the Error signal derivative. The set rules are defined as: ENG, ENP, Z, EPP, EPG.

| Inputs | Range in mm |
|---|---|
| Position in X axis | [-150, 150] |
| Error Signal Derivative | [-600, 600] |

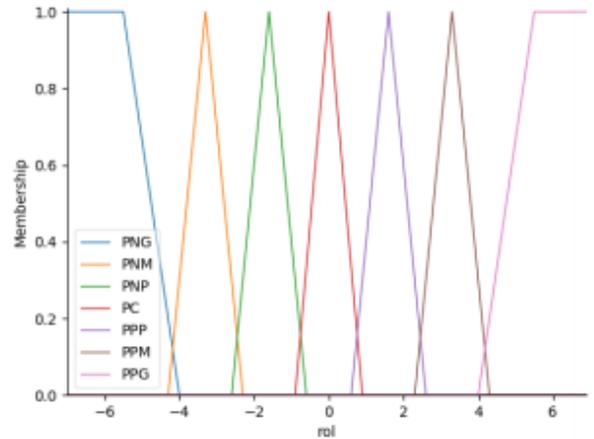

**Figure 11: Output with its Set of rules for Fuzzy 2.**

- **Fuzzy 3**: As input values we used the position of the ball in the X axis and the Error signal derivative. The set rules are defined as: NG, NP, Z, PP, PG.

| Inputs | Range in mm |
|---|---|
| Position in X axis | [-150, 150] |
| Error Signal Derivative | [-150, 150] |

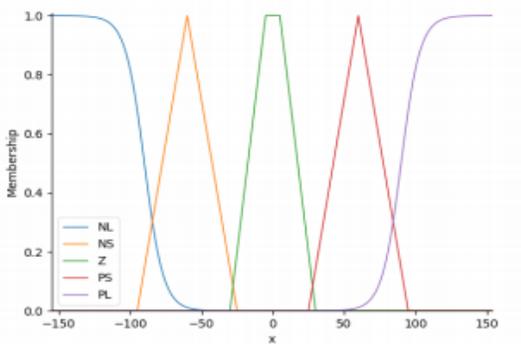

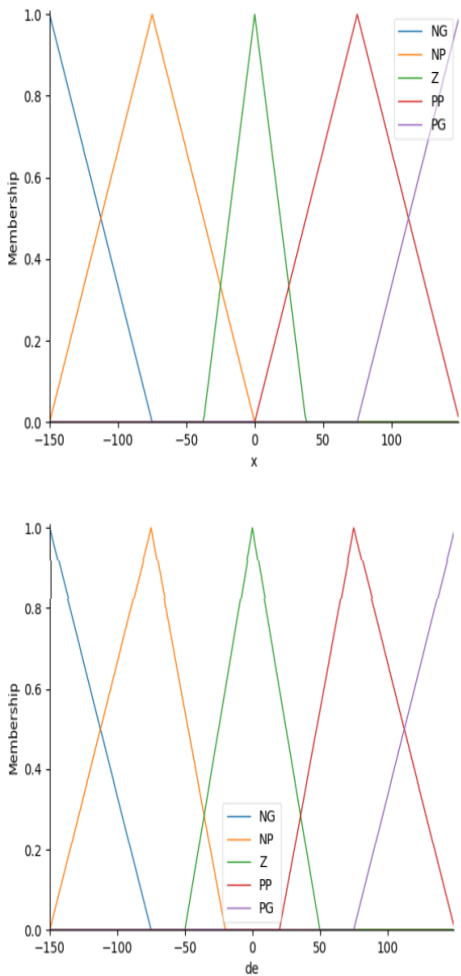

**Figure 12: Inputs with their Set of rules for Fuzzy 3.**

As output, only Roll parameter was considered and the set of rules are defined as: NG, NP, PC, PP, PG.

| Outputs | Range in degrees |
|---------|------------------|
| Roll    | [-4, 4]          |

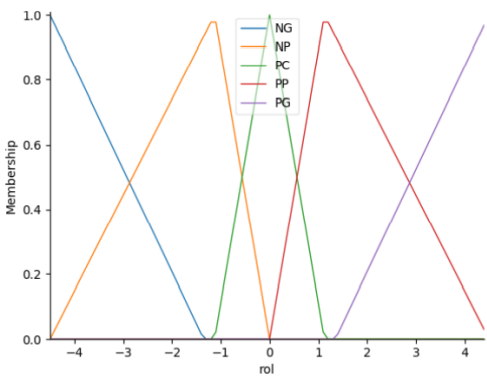

**Figure 13: Output with its Set of rules for Fuzzy 3.**

- **Fuzzy PD**: It is called PD due to the proportional and derivative parameters besides the regular inputs, which they are the position of the ball in the X axis and the Error signal derivative. The set rules are defined as: ENG, ENP, Z, EPP, EPG.

| Inputs                  | Range in mm   |
|-------------------------|---------------|
| Position in X axis      | [-150, 150]   |
| Error Signal Derivative | [-600, 600]   |

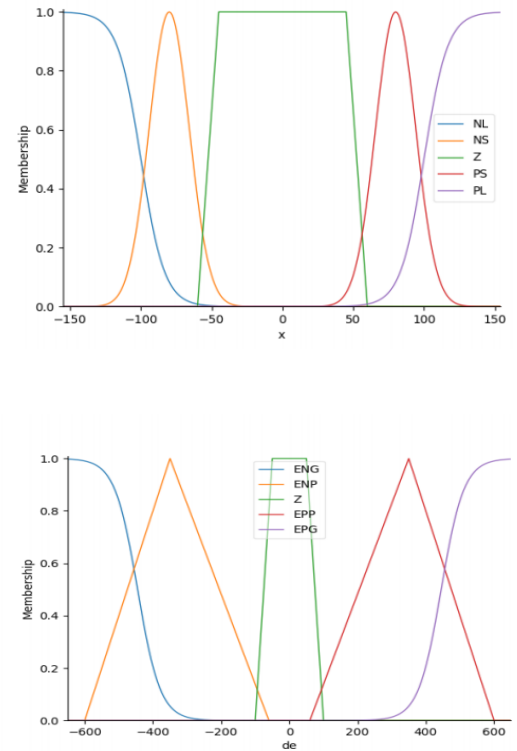

**Figure 14: Inputs with their Set of rules for Fuzzy PD.**

As output, only Roll parameter was considered and the set of rules are defined as: PNG, PNM, PNP, PC, PPP, PPM, PPG.

| Outputs | Range in degrees |
|---------|------------------|
| Roll    | [-5, 5]          |

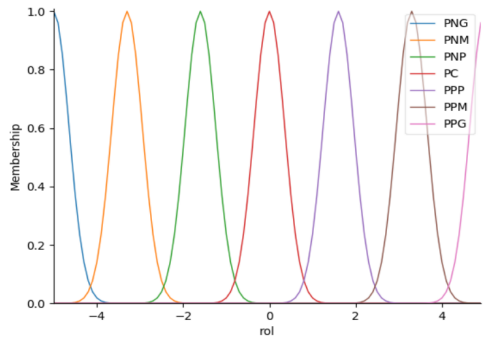

**Figure 15: Output with its Set of rules for Fuzzy PD.**

Once the set of rules, inputs, outputs, and their type of waves (sine, square, triangle form) have been defined for each Fuzzy test, we have used the Python package **skfuzzy 0.02.** to simulate and get an accurate behaviour of our systems..

- **Fuzzy 1:**

Each try, the ball was placed in one of the platform's corners, for this model, every simulation showed that the ball reached the desired area but stabilization time was not as fast as it was expected.

Its average oscillation range is within 75 mm around the center and stabilization time was never reached.

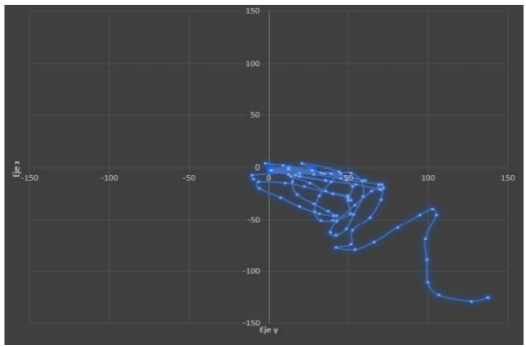

**Figure 16: Fuzzy 1 simulation.**

- **Fuzzy 2:**

This model lacks stability and precision, every simulation showed that the ball oscillates in a non-desired area before getting closer to the center of the platform. Once the ball reached this area, it was almost impossible to keep it there.

Its average oscillation range is within 40 mm around the center and stabilization time was never reached.

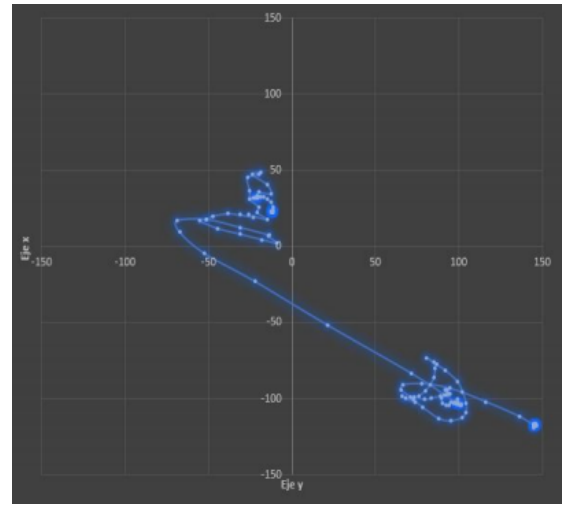

**Figure 17: Fuzzy 2 simulation.**

- **Fuzzy 3:**

Every simulation never reached neither stability nor accuracy. The ball once was close to the center, the technique sent it away and so on, this was a repetitive pattern.

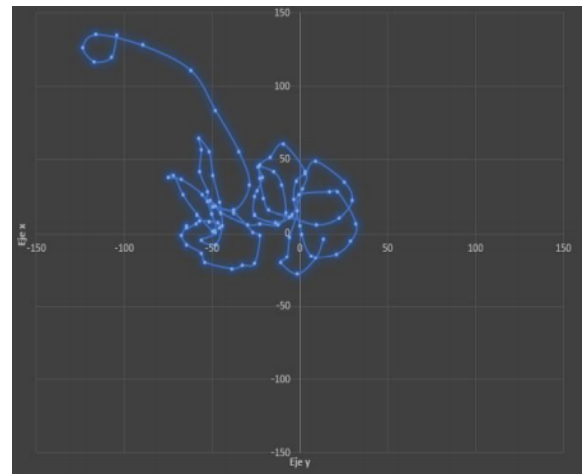

**Figure 18: Fuzzy 3 simulation.**

- **Fuzzy PD:**

Due to its proportional and derivative parameters, it allowed us to obtain a better control over the ball. Its oscillation range in the center is really close and stabilization time is fast enough to consider this model as our definitive one.

Its average oscillation range is within 10 mm around the center and stabilization time is 40 seconds since the ball is placed on the platform.

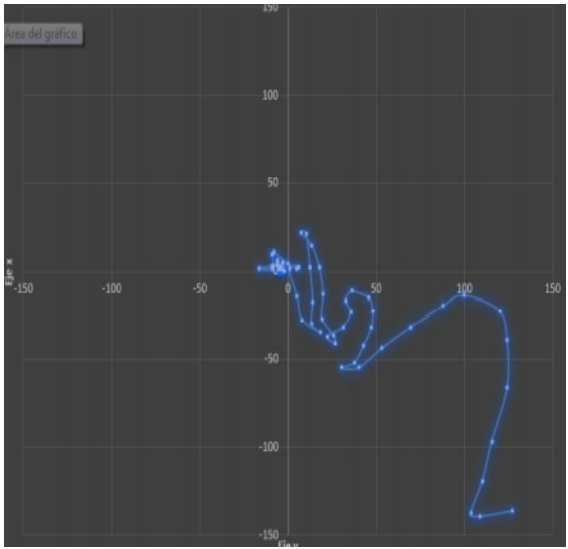

**Figure 19: Fuzzy PD simulation.**

After we chose our model, we run another simulation where we can appreciate how fuzzy works and describe in its own way the platform:

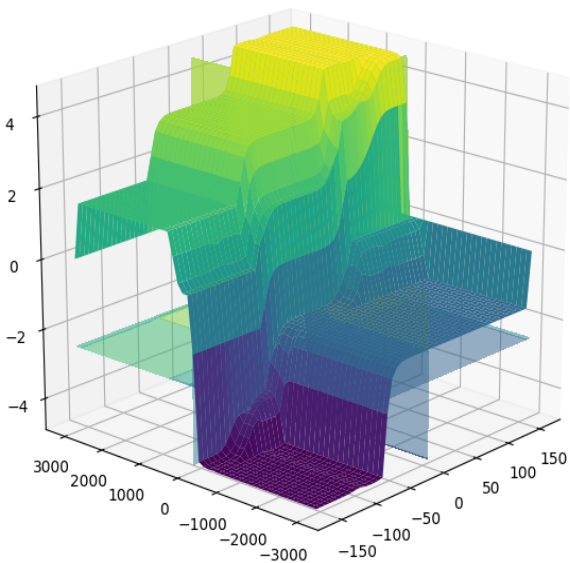

**Figure 20: Graphic Representation of Fuzzy PD model.**

Tuning the parameters affected in how fast the system and the stability of the platform during the simulation was fundamental.

## VI. CONCLUSIONS

The design of the Stewart Platform is directly related to the difficulty of its control. If the number of joints in the platform is greater than 3, the platform controller needs to be more complex.

The use of Python allows integrating a greater number of digital tools, at the same time offering a greater variety of solutions for the same problem.

A controlled environment was necessary to use artificial vision as part of the control method. The interaction between the ball, camera and platform was a high-difficulty target due to reflection of the light emitted from the platform. This problem was solved when the platform color was changed to black.

A modern technique control allowed us to achieve our goal sooner than It was expected. There are many techniques like neural networks or AI or even Deep machine learning that can help to solve and control the system.

A greater goal to accomplish is to use a modern technique from the beginning (develop the kinetic of the platform and how the servo motors move). Either way, reaching this goal is very complex due to very little data related to the movement of the platform and so many configurations or designs about the same system.

We might as well create groups of platforms with the same characteristics or features and start to obtain data to create a more sophisticated control. All code was written using Python coding language.